\documentclass{article}

\usepackage{arxiv}

\usepackage[utf8]{inputenc} 
\usepackage[T1]{fontenc}    
\usepackage{hyperref}       
\usepackage{url}            
\usepackage{booktabs}       
\usepackage{amsfonts}       
\usepackage{nicefrac}       
\usepackage{microtype}      
\usepackage{lipsum}
\usepackage{graphicx}
\graphicspath{ {./images/} }

\usepackage{amssymb}
\usepackage{amsmath}

\usepackage{booktabs}
\usepackage{float}
\usepackage[section]{placeins}
\usepackage{amsmath,amsfonts}
\usepackage{algorithmic}
\usepackage{algorithm}
\usepackage{array}
\usepackage[caption=false,font=normalsize,labelfont=sf,textfont=sf]{subfig}
\usepackage{textcomp}
\usepackage{stfloats}
\usepackage{url}
\usepackage{verbatim}
\usepackage{graphicx}
\usepackage{hyperref}
\usepackage{longtable}

\title{Probabilistic Physics-Aware Machine Learning Predictions of Electric Truck Energy Consumption with Field Data} 

\author{
 Hannes Nilsson \\
  Volvo Technology AB\\
  Gothenburg, Sweden\\
  Department of Electric Engineering \\
  Chalmers University of Technology \\
  Gothenburg, Sweden \\
  \texttt{hannesni@chalmers.se}
  \And
 Rafael Basso \\
  Volvo Technology AB\\
  Gothenburg, Sweden\\
  \texttt{rafael.basso@volvo.com}
  \And
 Balázs Kulcsár  \\
  Department of Electric Engineering \\
  Chalmers University of Technology \\
  Gothenburg, Sweden \\
  \texttt{kulcsar@chalmers.se}
  \And
 Morteza Haghir Chehreghani  \\
  Department of Computer Science and Engineering \\
  Chalmers University of Technology and University of Gothenburg \\
  Gothenburg, Sweden \\
  \texttt{morteza.chehreghani@chalmers.se}
}

\begin{document}

\maketitle

\begin{abstract}
\noindent In this work, we incorporate first principle physics into the construction of data-driven methods by considering a model that accounts for the different sources of energy losses during vehicle operations. Our results show that Bayesian linear regression based on this physics-aware model can improve the reliability of the expected energy consumption, as compared with standard linear regression. Further, it is shown that more complex machine learning models such as neural networks and gradient boosted regression trees, based on the same physical model, can further improve the accuracy in energy forecasting and significantly outperform standard versions of the same machine learning models. In addition to point predictions of the energy consumption, we develop a framework for estimating the corresponding uncertainty in the form of predicted standard deviation. Our results show that all of the models learn to estimate the uncertainty reasonably well.
\end{abstract}




\section{Introduction}
The global transition toward sustainable transportation has accelerated the adoption of electric vehicles (EVs), including electric trucks, which are expected to play a significant role in decarbonizing the freight and logistics sector \cite{Ahmed2025}. To aid this transition, accurate energy prediction for electric trucks is critical to ensure optimal route planning, battery management, charging infrastructure utilization, and overall operational efficiency. However, the electric energy consumption is influenced by multiple interacting factors, including vehicle dynamics, driving behavior, terrain, weather conditions, and traffic patterns. Traditional deterministic models, relying on physical relationships and factors predetermined in experimental setups often fail to capture this complexity adequately. Under varying real-world conditions, accurate estimation of inputs make up a great challenge for such methods, which can greatly impact their performance. 

In recent years, machine learning (ML) has emerged as a powerful tool for modeling intricate relationships in complex systems. ML-based energy prediction models have demonstrated superior accuracy over physics-based approaches by learning directly from historical driving and sensor data. In addition to improving prediction accuracy, data-driven methods offer natural ways to estimate the uncertainty in the ML models' predictions. Due to uncertainty of input parameters and intrinsic complexity of the input to output mapping, perfect predictions are out of reach. For optimal mission planning, then, trustworthy estimation of the expected deviation from the trucks' most probable energy consumption is crucial. Nevertheless, most existing ML approaches focus solely on point predictions, overlooking the inherent uncertainty associated with energy consumption forecasts. This omission can be problematic in high-stakes applications such as long-haul freight, where underestimating energy needs may lead to range anxiety or delivery failures, while overestimation may result in inefficient use of charging infrastructure and large unnecessary costs for over-sized battery.

To address these limitations, this study investigates the application of machine learning models for predicting the energy consumption of electric trucks, with a specific focus on uncertainty estimation. By incorporating uncertainty quantification, the models not only provide accurate predictions but also communicate the deviation one can expect associated with each estimate. This is essential for risk-aware decision-making, enabling fleet operators and route planners, regardless of whether they are human or autonomous agents, to make informed decisions under uncertainty. Through experiments on extensive real-world driving datasets, this work aims to advance the state-of-the-art in predictive modeling for electric truck energy consumption, ultimately contributing to more reliable and efficient electric freight operations.

\subsection{Related Work}
The literature study in \cite{zhu2024} provides a thorough up-to-date overview of the current techniques and solutions. According to this review, the previous works can be broadly categorized into three different approaches: Empirical, physics-based and data-driven methods. We adopt the same grouping here.

Physics-based methods make use of Newton's second law of motion, as it applies to vehicles. For a good overview of this approach, see \cite{Guzzella}. This way, the dynamics of a vehicle (not considering any applied manual braking) can be expressed in terms of the instantaneous wheel traction force required to maintain or change the vehicle speed:
\begin{equation} \scalebox{1.0}{%
    $
    f = ma + mg \sin \theta +mg C_r \cos \theta + \frac{1}{2} C_d A \rho v^2.
    $}
    \label{phys_eq_force}
\end{equation}
Here, $ma$ denotes the net acceleration force, where $m$ is the vehicle mass and $a$ is its acceleration. The gravitational force due to road inclination is given by $mg \sin \theta$, with $g$ being the gravitational acceleration and $\theta$ the road slope angle. The rolling resistance force is modeled by $mgC_r \cos \theta$, where $C_r$ represents the rolling resistance coefficient. Finally, the aerodynamic drag force is expressed as $\frac{1}{2} \rho C_d A v^2$, where $\rho$ is the air density, $C_d$ the drag coefficient, A the frontal area, and $v$ the vehicle velocity. Taking the powertrain efficiency $\eta$ into account, and considering short enough time steps where these parameters can be assumed to be constant, integrating the force over the distance $d$ gives the following equation for energy consumption:
\begin{equation} \scalebox{1.0}{%
    $
    e = \frac{1}{\eta} \left( mad + mgd \sin \theta + mg C_r d \cos \theta + \frac{1}{2} C_d A \rho d \left(\frac{v_i^2 + v_f^2}{2}\right) \right)
    $}
    \label{phys_eq_energy},
\end{equation}
where $v_i$ and $v_f$ are the initial and final velocity over the short time step.

The coefficients of rolling resistance and air drag are complex parameters that depend on both vehicle-specific and environmental factors. Furthermore, energy consumption also depends significantly on the powertrain efficiency which is hard to estimate and vary with the applied torque and current speed. If one wishes to model all these dynamic effects completely, there is an intractable number of parameters and microscopic effects to consider. In absence of a perfect model, numerous attempts have been made to describe the dynamics sufficiently accurate, at levels of varying detail \cite{BASSO2019}, \cite{FIORI2016}, \cite{YUAN2017}, \cite{YANG2014}. These physics-based approaches suffer from two main issues when it comes to accurate energy consumption prediction: Computational complexity and reliance on accurate input prediction. A more detailed physical model is generally better in terms of accuracy, as it is able to account for more effects. At the same time, this requires more computations to be done which may put pressure on both hardware requirements and time limitations.

Often inspired by the physical formula above, empirical models assume that the energy consumption follows simple functional relationships. The very simplest ones consider it as a linear function of the driving distance, such as in \cite{Li2019} and \cite{xylia2017}. More advanced empirical methods derive relatively simple parametric functions of one or a few variables. Examples of such models can be found in \cite{Fetene2017}, \cite{Wu2015}, \cite{Ji2022} and \cite{zhao2019}. Apart from driving distance, common parameters to consider are velocity, acceleration, mass, air temperature and travel time, all of which have a direct or indirect effect on the electric energy consumption. Although overcoming the computational limitations of physics-based modeling, and to some degree avoiding having to predict many of the parameters on a detailed level, empirical models struggle to capture the full complexity that electric energy consumption constitutes.

Before describing the models categorized as data-driven, we should make clear that most, if not all of the empirical and physics-based models proposed in the literature use data for calibration. Physics-based models rely on parameters such as the coefficients of rolling resistance, air drag, and powertrain efficiency, which are often estimated from data collected during experiments. Similarly the parameters that empirical models depend on are often tuned with experimental data. In contrast to empirical or physics-based models calibrated using driving data, we consider data-driven methods as ones which learn relationships between the independent and dependent variables that are not explicitly constructed beforehand, using real-world data collected on the road. This distinction will help better understand the differences between some of the methods we propose in this paper, which we call physics-aware machine learning methods and in a sense lie on a spectrum between physics-based and completely data-driven. 

Similar to empirical models, data-driven ones often take inspiration from Equation \ref{phys_eq_energy} in some way. \cite{zhu2024} uses the known physics of vehicle dynamics for feature engineering, aiming to construct informative features which correspond to different parts of \ref{phys_eq_energy} from measured vehicle data. The resulting features are then fed through machine learning methods (neural networks and regression tree ensembles), trained to predict the resulting energy consumption. The need for such expressive models arises from the fact that this work aims to predict the energy consumption of a trip directly, using averages, variances and quantiles of the constructed features over the whole distance traveled. Additionally, these types of machine learning methods are also able to account for effects that are not expressed in the physical formula, due to their complexity, such as the effect that temperature and wetness of the road surface have on rolling resistance, and through that on the energy consumption. Similar trip-level methods for predicting energy consumption with machine learning and high-level aggregated features have been proposed in \cite{LI2021}, \cite{LI2021_2}, \cite{ZHANG2020}. An interesting aspect of \cite{zhu2024} that distinguishes it from the other similar works is that they set out to predict the uncertainty in the machine learning predictions, through quantiles, in addition to the most likely point estimates. We set out to investigate prediction uncertainty in this paper too, albeit with a somewhat different approach. 

One issue of the high-level methods discussed above, which directly predict the energy consumption of a whole trip at once is that they risk missing detailed information about the varying conditions along the route. Although these methods build features that are meant to capture the relevant attributes along the path, as per Equation \ref{phys_eq_force}, such as average road inclination, acceleration and squared speed, the averaging process throws away a lot of detailed information. Some of the high-level methods account for this to some extent by including variability in the features, in the form of quantiles, variance or both, which appears to improve prediction accuracy \cite{zhu2024}. Even with features capturing some kind of variability, however, information is lost in the process and the fact that accuracy improves with such features hints that a more detailed consideration of the situation along the route is advantageous. In contrast to these high-level methods, \cite{DeCauwer2027} takes a more low-level approach by setting out to learn the coefficients of a model derived from a physical formula for energy consumption similar to \ref{phys_eq_energy}, using multiple linear regression. With such strong dependence on the physical equation, one might consider their approach as a physics-based one. However, as the coefficients are learned from large sets of data collected on-the-road during real-world driving, the authors call the method data-driven. This agrees with the distinction we make between the two different practices too, although we acknowledge that this approach lies somewhere in-between. 

Another interesting low-level approach is taken by \cite{Thorgeirsson2021}. They use a two-scale approach, where energy predictions are first made on high-frequency data of 10Hz from logged signals, whereafter a second model aims to correct the first model's prediction using averaged features over a whole road segment. A road segment is defined as the part of the road in-between two intersections, usually in the order of a few hundred meters, so the corresponding predictions are much shorter than full-trip predictions. Using this setup, this work considers both multiple linear regression and neural networks for the energy prediction problem. A major difference compared to the situation we study in this work is that they focus on federated learning---an approach where multiple vehicles learn together and update a collective model, which requires constant streaming of data between the vehicles, often through a central node. Like \cite{zhu2024}, this study is also interested in the predictive uncertainty of the learned models. The difference here is that this paper considers the predictive variance as a measure of uncertainty, rather than quantiles---which is the same approach we take in this work.

Lastly, \cite{Litjens2025} introduces methods that are similar to \cite{Thorgeirsson2021} in that they use machine learning on high-frequency data to predict both the mean and variance on short segments that are added up to form full trip predictions. One major difference, however, is that this work focuses on LSTM architectures which are more complicated than the linear regression and multi-layer perceptrons model used in \cite{Thorgeirsson2021}.

\subsection{Contributions}
In light of the previous works described above, we see a need for bringing the physics-based and data-driven approaches closer together, in order to leverage the relative strengths of each framework. On the one hand, machine learning methods are capable of identifying relationships in the data that are hard to model physically, and their statistical nature offer ways to estimate the predictive uncertainty in addition to most likely point estimates. On the other hand, physics-based methods derived from first principles do not require a large amount of data, provide interpretable predictions, and can generalize to conditions which have not been seen before. For instance, purely data-driven approaches with high-level data were proven successful in \cite{zhu2024}, where the models had access to a huge dataset from over 90,000 taxi trips, all driven in the city of Beijing. In practice, and especially for long-haul truck applications which is a target for this work, one cannot expect to have access to such amount of data, and must also consider a much broader set of driving conditions. In \cite{zhu2024}, they somewhat guide their ML models to respect the physical relations relevant for energy consumption of electric vehicles through the proposed feature construction process. Our work takes it one step further and builds this physical information into the models themselves. The main contributions of work are three-fold:
\begin{enumerate}
    \item We develop novel physics-aware machine learning models that explicitly incorporate the underlying physical phenomena leading to energy losses. This has been explicitly called for in previous work \cite{Litjens2025}.
    \item We propose new ways of estimating and calibrating the uncertainty in energy consumption prediction for electric vehicles using machine learning methods that leverage detailed information from high-frequency data, without assuming independence between connected road segments which is common yet unrealistic. To the best of our knowledge, this is also the first work to investigate variance estimation from gradient boosted regression trees for this problem.
    \item We demonstrate the practical applicability and robustness of our approach by evaluating the models on extensive real-world data collected directly from electric trucks operating on the road, successfully bridging the gap between theoretical vehicle dynamics and complex driving conditions, without relying on simulations.
\end{enumerate}

In the following section, we will introduce a set of physics-aware machine learning models, which are built around Equation \ref{phys_eq_energy} while still leveraging large-scale data. The data is collected on the road to learn functions of energy consumption that are both tailored to the specific vehicle producing the data, and account for variations in the driving conditions. Our study focuses on three types of machine learning models: Linear, Neural Networks, and Tree Ensembles. For each class of models, we investigate methods to quantify uncertainty in the form of their predictive variance. Furthermore, we evaluate both standard versions of the models and their Bayesian variants. Bayesian methods allow for incorporating information into the models beforehand, by means of a prior distribution. This can provide the benefit of learning quicker from less data, and make the models less sensitive to noise. Furthermore, it can help making the models more interpretable, as the parameters of the models stay closer to their anticipated values, specified as prior means. In \cite{DeCauwer2027}, it was observed that while a linear model can find coefficients that work fairly well for energy consumption prediction, the parameters may not necessarily correspond to the physical phenomena they represent. In their work, the multiple linear regression models tend to set the coefficient of air drag to very small values---on the order of $10^{-5}$---while the true air drag coefficients of passenger cars typically lie between $0.2$ and $0.4$. This was one of the observation that led us to consider Bayesian multi-variate linear regression for the same problem as an alternative in this study.

\section{Energy prediction models}
The goal of this work is to develop machine learning based energy prediciton models that strike a balance between interpretability, accuracy, computational efficiency and the ability to generalize to new routes and variations in driving conditions. To this end, our methods use time-series data of 1Hz, which is smoothed with a non-overlapping averaging window of 10 samples. This window size appears to strike a good balance between averaging out the inherent noise in the measurements of the sensors, while maintaining detailed information about the driving scenario. The data is collected from four electric trucks out of Volvo's internal fleet of test vehicles---two trucks for long-haul transport and two for regional haulage---fitted with all necessary equipment for the data collection

In addition to the measured data, we append the time-series data with data regarding the driving environment and weather. The weather-related features we consider are fetched from meteostat \cite{LamprechtMeteostatPython} and reported on an hourly basis. Moreover, the wind data is used in combination with the vehicle speed and driving direction to calculate the speed of the vehicles relative to wind, which is of interest for estimating air drag as per Equation \ref{phys_eq_energy}. A summary of all the features used to train the machine learning models and make the predictions is reported in Table \ref{tab:features}.

\begin{table}[t]
\centering
\begin{tabular}{ll}
\hline
\textbf{Features for linear models} & \textbf{Additional features} \\
\hline
Mass ($m$) & Drive direction \\
Acceleration ($a$) & Precipitation \\ 
Distance driven ($d$) & Average vehicle speed \\
Road inclination ($\theta$) & Temperature \\
Vehicle speed relative to wind ($v$) & Humidity \\
Gravitational acceleration ($g$) & Air pressure \\
Air density ($\rho$) & Average wind direction \\
& Average wind speed \\
& Wind peak gust \\
\hline
\end{tabular}
\caption{Overview of measured, fetched and derived signals used by the different machine learning models.}
\label{tab:features}
\end{table}

As discussed above, developing models that predict energy consumption on a low level from high-frequency data is an attempt to make interpretable and generalizable models, where the known physics from Equation \ref{phys_eq_energy} applies. In the end, however, it is accurate energy consumption on a full-trip level we are after. Hence, with our approach, we let the models predict energy consumption on the short segments coming from the high-frequency data, and add up the individual predictions to make the full-trip predictions. It is these resulting full-trip predictions that we then evaluate and compare the models' performances on. The aspects we are interested in is each model's ability to capture the predictive distribution of future energy consumption, in the form of its mean and standard deviation. 

Consequently, the models are designed to predict both the mean $\mu_{l}$ and variance $\sigma^2_{l}$ of the energy consumption on each short segment, or link $l$ coming from the collected and smoothed data points. If we make the assumption that the energy consumption on all links that make up a trip are independent Gaussian random variables, this would allows us to add up all individual link predictions up to obtain the predictions $\mu_\text{trip}$ and $\sigma^2_\text{trip}$ on the full trip consisting of $L$ links:
\begin{equation} \scalebox{1.0}{%
    $
    \mu_\text{trip} = \sum_{l=1}^{L} \mu_l,
    \quad
    \sigma^2_\text{trip} = \sum_{l=1}^{L} \sigma^2_l
    $}\label{eq_mean_trip}.
\end{equation}
However, we acknowledge that this may not be entirely true. In particular, it is likely that there is a strong correlation between adjacent links, where deviations from the expected energy consumption may be caused by the same road conditions. For sums of dependent random variables, one must consider the covariance between them. For example, consider the case of summing up two dependent random variables $A$ and $B$, where we have the following:
\begin{equation} \scalebox{1.0}{%
    $
    \text{Var}[A+B] = \text{Var}[A] + \text{Var}[B] + 2\text{Cov}[A,B].
    $}
\end{equation}
In our case, where we sum up $L$ links there would be $L$ choose $2$ covariances to account for. At the same time, it is reasonable to assume that two links that are far apart, for instance the first and the last link behave more like independent random variables with low covariance. For the long trips we consider in this work, calculating the covariance between links of varying lengths of separation from $0$ to $L$ would be cumbersome. Instead, we choose to account for these additional terms of the variance in trip-level energy consumption by learning a covariance factor $c_\text{cov}$ multiplied by the number of links in the trip $L$, representing the average covariance between a link and all other links in a trip. 

Furthermore, previous work suggest that the variance learned by machine learning methods have a tendency to be poorly calibrated \cite{guo2017}, \cite{Levi2022}. \cite{kuleshov2018} proposes to use isotonic regression of the predicted variance. This method fits a non-decreasing function of the predicted variance with the aim of scaling it to empirically match the observations from a set of validation data points. In this work, we consider a simple form of isotonic regression, which fits a single multiplicative factor of the predicted variance on a link level. We apply this method of calibration to all the models before accounting for the covariances between individual predictions when summing up the link variances over full routes. 

Hence, for each model, we consider all link predictions in the complete set of validation routes and fit a multiplier $c_l$ so that for approximately 68\% of these links, $e_l$ is within the range $\mu_l \pm c_l \sigma_l$. After this, we will set out to find covariance factor $c_\text{cov}$, which is the last piece required for calculating the final variance prediction on a trip level:
\begin{equation} \scalebox{1.0}{%
    $
    \sigma^2_\text{trip} = \sum_{l=1}^{L} c_l \sigma^2_l + c_\text{cov}L
    $}\label{eq_var_cov_trip}.
\end{equation}
Before describing how we determine the factor $c_\text{cov}$, we will outline how the data from the trips is divided for learning and evaluation. Since one of the benefits with data-driven methods is to learn individualized functions of energy consumption---for instance learning to account for the current wear-and-tear of the tires which has an affect on the rolling resistance---individual machine learning models are implemented for each one of the four trucks. For each truck, its set of routes with corresponding time-series data is divided into three disjoint sets, with 50\% of the routes used for training machine learning models, 25\% for validation and the last 25\% for evaluating and comparing the performance of all models.

The first step towards obtaining machine learning model for predicting $\mu_\text{trip}$ and $\sigma^2_\text{trip}$ is then to build models that can predict $\mu_l$ and $\sigma^2_l$ for the individual road segment we have data for. This process uses the training set for each truck and differs depending on the specific machine learning models used. We will describe the process for the various models below. With such models in place, we first apply the models to predict $\mu_\text{trip}$ and $\sigma^2_\text{trip}$ on the set of validation routes using Equations \ref{eq_mean_trip} and \ref{eq_var_cov_trip} with $c = 0$, hence not accounting for any covariance between the links. By our Gaussian assumption, for $68$\% of the routes the true measured energy consumption should should fall within the mean plus/minus one standard deviation. Therefore, we increase the value of $c$ and recalculate $\sigma^2_\text{trip}$ until this condition holds for approximately $68$\% of the validation routes. Since we have a finite number of routes we will not achieve exactly $68$\%. We do the same for each of the four vehicles individually. 

It should be noted that for the methods to provide true future predictions, and to be applicable in practice, we would also be required to make predictions of the variables listen in Table \ref{tab:features}. However, our main objective in this work is to compare different machine learning approaches for this task, without the results relying on the accuracy of model inputs. Therefore, we assume we can model the inputs equally accurate for the future predictions as for the previous trips, and use the historical measured data for training and testing alike. This is a common approach in the previous works on the topic as well.

\subsection{Linear models}
Since Equation \ref{phys_eq_energy} displays a linear function in combinations of variables that are part of the features we have access to, our first and simplest machine learning models are based on multiple linear regression. Here, the aim is to learn the multiplicative factors of the different terms of the function, in order to obtain a digital twin model that represents the specific truck. In other words, only a subset of all features are used for the linear models. For the first and second term, this multiplicative factor corresponds to the reciprocal of the powertrain efficiency $\eta$, and can be combined into one. For the third term, this multiplicative factor represents the coefficient of rolling resistance divided by $\eta$ and for the forth it is the air drag times frontal surface area divided by four times $\eta$. This can be seen directly with a simple rearrangement of the equation:
\begin{equation} \scalebox{1.0}{%
    $
    e = \frac{1}{\eta}(mad + mgd \sin \theta) + \frac{C_r}{\eta}mg d \cos \theta + \frac{C_d A}{4 \eta} \rho d \left(v_i^2 + v_f^2\right)
    $}\label{phys_eq_linreg}.
\end{equation}

\subsubsection{Multiple linear regression} Multiple linear regression (MLR) is an extension of regular linear regression with ordinary least squares. However, instead of fitting the line that minimizes the sum of squared residuals, MLR does the same with a hyperplane in $D$ dimensions, where $D$ is the number of linear features, which is three in our case. As MLR finds the best fit hyperplane to explain the measured energy consumption based on the linear features, they consider the multiplicative factors as constants. Since we know that these factors carry physical meaning in the form of rolling resistance, air drag and powertrain efficiency, we know that this is not true in reality, and that they are in fact varying with the driving conditions. Therefore, what the multiple linear regression models learn can be seen as the average values of these multiplicative factors, which have some natural level of fluctuations around them.

\subsubsection{Bayesian multivariate linear regression} Due to the inherent fluctuations of the multiplicative factors displayed in \ref{phys_eq_linreg}, it makes sense to model these parameters as random variables, rather than constants. In light of this, we investigate an alternative to multiple linear regression, called Bayesian multivariate linear regression (BMLR) \cite{Wiley1992}. 

The Bayesian school of statistics is based upon Bayes' rule, which is a theorem that updates the probability of a hypothesis, based on new evidence. For the problem at our hands, the hypothesis corresponds to the values of the linear regression coefficients, and the evidence comes in the form observed time-series data. More specifically, we are interested in finding the posterior distribution of the unknown coefficients $\mathbf{w} = \Big[\frac{1}{\eta}, \frac{C_r}{\eta}, \frac{C_d A}{4 \eta} \Big]^T$ from the linear Equation \ref{phys_eq_linreg}, given the observed values for energy consumption $\mathbf{t}$ over the short segments. This posterior probability is expressed using Bayes' rule as:
\begin{equation} \scalebox{1.0}{%
    $
    p(\mathbf{w}|\mathbf{t}) = \frac{p(\mathbf{w}) p(\mathbf{t}|\mathbf{w})}{p(\mathbf{t})}
    $}\label{bayes_rule}.
\end{equation}
Here, $p(\mathbf{w})$ is the initial probability of certain parameter values $\mathbf{w}$ being the true ones before any new evidence is considered. We denote the mean vector of initial parameter values as $\mathbf{m}_0 = \Big[\left(\frac{1}{\eta}\right)^\text{ref}, \left(\frac{C_r}{\eta}\right)^\text{ref}, \left(\frac{C_d A}{4 \eta}\right)^\text{ref} \Big]^T$, which come from controlled experiments performed for a specific truck models and tires. $p(\mathbf{t})$ gives the overall probability of the measured energy consumption $\mathbf{t}$ and $p(\mathbf{t}|\mathbf{w})$ is the probability of observing these measurements given that $\mathbf{w}$ are the true coefficients. In \ref{appendix:bayesian_details}, we outline how to obtain $p(\mathbf{w}|\mathbf{t})$. The derivations are adapted from \cite{bishop2007} which contains more details.

\subsubsection{Uncertainty estimation} For the linear models, Bayesian and standard alike, we assume homoscedastic noise, which means that the variance of the error term in the models is constant across all levels of the independent variables, so that $\sigma^2_l$ is the same for all links. This way, we can estimate $\sigma^2_l$ in a straight-forward manner. After fitting the linear models to the links from the training routes, we run the models on the same data to obtain $\mu_l$ for all links. Then, we consider $\sigma^2_l$ as the sample variance of the residuals $e_l - \mu_l$, where $e_l$ is the measured energy consumption on link $l$.

Although the linear models are based on the known physics of the problem as described by Equation \ref{phys_eq_energy}, we cannot expect them to account for all the physical effects contributing to energy consumption. In particular, many of the factors that appear in the equation are either non-stationary by nature, or hard to measure accurately and reliably. We will address these issues in more detail in Section \ref{sec:discussion}, but it is important to understand that what these linear models learn from the data may not exactly correspond to the quantities that the equation suggest, since they will also correct for potential biases and errors in the measured data features. Furthermore, neither the powertrain efficiency $\eta$ nor the coefficient of rolling resistance $C_r$ are constant values. $\eta$ depends on the current gear, speed and torque of the vehicle, while $C_r$ vary depending on things like tire temperature and road surface. In light of this, what the linear models are able to learn are average values for the multiplicative factors, under the driving conditions of the training data, that best captures energy consumption prediction.

\subsection{Neural networks}
As an attempt to be able to make even more accurate future predictions, we develop models based on more flexible machine learning, namely artificial neural networks. These models are capable of accounting for more complicated phenomena that are hard to model physically, and are not restricted to consider constant average values for the multiplicative factors.

\subsubsection{Physics-aware neural networks}
In practice, neural networks are commonly used as universal black-box function approximators where they are directly applied to predict the quantity of interest, based on previous data. In our case, that would mean that the networks are trained to output estimates of energy consumption directly. However, there are many issues with this approach, such as overfitting poor generalization and interpretability \cite{ROHLFS2025}. Also, since we already have a validated physical equation for the problem, it would be unwise not to use it when going from linear models to more advanced machine learning methods. Therefore, we develop what we call physics-aware neural networks that respects the known physics while flexibly learns from the data at hand.

We incorporate the physical information by letting the neural networks output three quantities corresponding to correction terms to be added to the reference values of the multiplicative factors $\mathbf{m}_0$. After this process, the estimated energy consumption is calculated with Equation \ref{phys_eq_linreg} just like for linear regression. The features we feed into the neural networks are the ones presented in Table \ref{tab:features}.

\begin{figure}[!t]
\centering
\includegraphics[width=\textwidth]{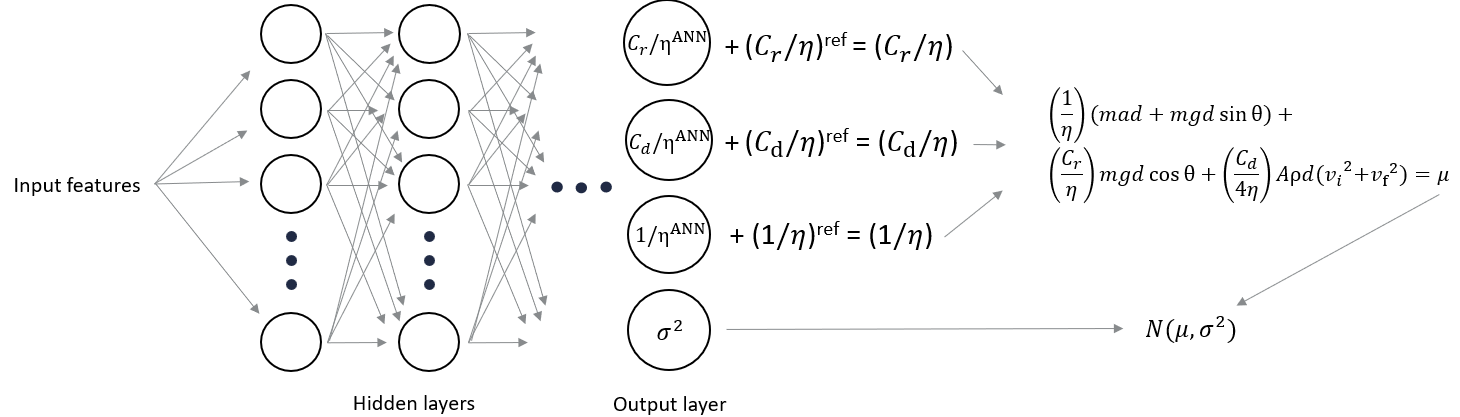}
\caption{Our physics-aware neural network architecture for energy consumption prediction.}
\label{fig:nn}
\end{figure}

A schematic view of the physics-aware neural network architecture is shown in Figure \ref{fig:nn}. As this figure suggests, there is an additional forth output of the neural networks corresponding to the predicted variance. Since the networks are able to adapt the predictions to the circumstances described by the inputs, we can move beyond a homoscedastic variance assumption as in the linear case, and let the networks directly predict the variance from the input features. Therefore, the target data samples in the training sets are treated as samples from Gaussian distributions with expectation and variances. For the networks to learn to accurately predict these quantities, the networks are trained with Gaussian negative log-likelihood (GNLL) loss, rather than e.g. mean-squared error loss or some other popular loss function used in neural network training for point prediction tasks. The GNLL loss function $\mathcal{L}_\text{GNLL}$ is defined as:
\begin{equation} \scalebox{1.0}{%
    $
    \mathcal{L}_\text{GNLL} = \frac{1}{2} \sum_{l=1}^L \left[ \text{log} ( 2 \pi \sigma^2_l ) + \frac{(e_l - \mu_l)^2}{\sigma^2_l} \right],
    $}\label{eq_gnll}
\end{equation}
where $L$ is the number of links in the corresponding training set.

We acknowledge that our physics-aware neural networks, like standard neural networks, still act as black-box models that provide little insight into how they learn to correct the multiplicative factors. However, this is true also for the linear models, and here can similarly observe how much emphasis the networks give to the different sources of energy consumption. 

\subsubsection{Bayesian physics-aware neural networks} In addition to the physics-aware neural networks described above, we also implement a Bayesian version based on Bayesian neural networks (BNNs) \cite{MacKay1995}, \cite{LAMPINEN2001}, \cite{Wang2020}, \cite{Goan2020}. The overall architecture is identical in the two cases, and the models only differ in the way the networks are trained. In the Bayesian case, a probability distribution is placed over the learnable parameters $\mathbf{\omega}$, commonly referred to as the network's weights, which are then viewed as random variables. These weights are hidden or latent variables, for which we have no way of directly observing their true underlying values. However, Bayes rule \ref{bayes_rule} provides us with a way to reason about the posterior distribution of the weights $p(\mathbf{\omega}|\mathbf{X}, \mathbf{t})$, given the data we observe, with features $\mathbf{X}$ and targets $\mathbf{t}$. We will not go into detail here about the derivations and theoretical aspect of Bayesian neural networks. Instead, we refer the curious reader to the sources listed above.

In practice, the difference in training Bayesian neural networks, as opposed to the regular ones described earlier, is that the loss function will contain an additional term $\text{KL}\big(p(\mathbf{\omega})||q_\mathbf{\phi}(\mathbf{\omega})\big)$. The term comes from the Kullback-Leibler divergence between a pre-specified prior distribution $p(\mathbf{\omega})$ and $q_\mathbf{\phi}(\mathbf{\omega})$ which is a parameterized approximation to the posterior distribution of the weights $p(\mathbf{\omega}|\mathbf{X}, \mathbf{t})$. This measure is a way of quantifying the similarity of two probability distributions, and the incorporation of this additional term allows us to guide the learning process of the networks. 

Although widely successful for many tasks, neural networks do have their drawbacks, one of which being vulnerability to overfitting \cite{salman2019}. This phenomenon occurs when the neural network adapts too much to the training data it is presented with, to the extent that the predictions are distorted by random patterns and noise in the training set, which may negatively affect the model's ability to make accurate predictions on unseen data. In this regard, the Bayesian formulation of neural network training may have a regularizing effect, through an informative prior weight distribution $p(\mathbf{\omega})$. To this end, we consider Gaussian priors with mean $\mu_{\omega_0} = 0$ and with a small variance $\sigma^2_{\omega_0}$ for the network weights. By design, our physics-aware neural networks aim to learn corrections to the expected multiplicative factors of Equation \ref{phys_eq_linreg}, rather than the full values of the factors or energy consumption directly. Therefore, it makes sense to consider prior weight distributions with zero mean, and only when the data is convincing enough will the network update the weight values be large enough in absolute value so as to affect the predictions significantly. The trade-off between being able to learn from the data, while not overfitting to noise can be controlled by varying both the prior weight variance $\sigma^2_{\omega_0}$, and a multiplicative relative weight factor of the $\text{KL}$ loss term $\alpha$. We consider independent Gaussian distributions $q_\mathbf{\phi}(\mathbf{\omega})$ for all weights as well, which is a common assumption, so that the parameters $\mathbf{\phi}$ simply correspond to the posterior means $\mathbf{\mu}_{\omega_L}$ and variances $\mathbf{\sigma}^2_{\omega_L}$, which can be estimated through sampling during the training process. The full expression for the loss function of the Bayesian neural networks looks as follows:
\begin{equation}\scalebox{1.0}{%
$
\begin{aligned}
    \mathcal{L}_\text{BNN} = \mathcal{L}_\text{GNLL} + \alpha \mathcal{L}_\text{KL} = \sum_{l=1}^L \left( \frac{1}{2} \left[ \text{log} ( 2 \pi \sigma^2_l ) + \frac{(e_l - \mu_l)^2}{\sigma^2_l} \right] \right) + \alpha \left( \text{log} \frac{\sigma_{\omega_L}}{\sigma_{\omega_0}} + \frac{\sigma^2_{\omega_0} + (\mu_{\omega_0} - \mu_{\omega_L})^2}{2 \sigma^2_{\omega_L}} - \frac{1}{2} \right).
\end{aligned}
$}\label{eq_bnn_loss}
\end{equation}
Reframing neural network training as a Bayesian inference problem is not the only way to tackle the issue of overfitting. Another approach is to use weight decay, which refers to the process of adding an additional term to the loss function which penalizes large weight values \cite{weight_decay_1991}. For the standard neural networks, we apply L2 weight decay, which adds a regularization term consisting of the squared magnitude of the networks' weights multiplied by a factor $\lambda$, called the regularization parameter. By this addition, the final loss function of the neural networks becomes:
\begin{equation} \scalebox{1.0}{%
$
\begin{aligned}
    \mathcal{L}_\text{NN} = \mathcal{L}_\text{GNLL} + \lambda \mathcal{L}_\text{L2}
    = \sum_{l=1}^L \left( \frac{1}{2} \left[ \text{log} ( 2 \pi \sigma^2_l ) + \frac{(e_l - \mu_l)^2}{\sigma^2_l} \right] \right) + \lambda \left( \sum_{i=1}^\Omega \omega_i^2 \right),
    \end{aligned}
    $}\label{eq_nn_loss}
\end{equation}
where $\Omega$ is the number of weights in the network.

In addition to the respective regularization techniques of the two variations of neural networks, both are also trained using dropout \cite{srivastava2014}. This trick blocks the connections to a proportion of neurons each time we feed some input data through it. This way, no neuron becomes too specialized in predicting certain aspects of the data, and is another way of tackling overfitting. We also want to clarify that although the equations \ref{eq_bnn_loss} and \ref{eq_nn_loss} portray the total loss of a complete training set with $L$ data points, which are road segments in our case, the networks are trained with stochastic gradient descent using mini-batches. To implement and train the neural networks, PyTorch \cite{Pytorch2019} is used, with the built-in Adam optimizer \cite{kingma2017}.

\subsection{Gradient boosted regression trees}
Although neural networks have demonstrated impressive performance on many machine learning tasks, and have received a lot of attention lately, they are not always the one-and-only choice of model. They might be favored when it comes to generative artificial intelligence and problems involving perception, such as computer vision and natural language processing tasks. However, when it comes to predictions from tabular data, the family of models known as tree ensembles have been widely successful \cite{Borisov2022}, \cite{grinsztajn2022}, \cite{gorishniy2023}, \cite{nilsson2024}. An up-to-date overview of tree ensemble methods is provided by Blockeel et al. [2023], where several advantages are discussed over other techniques. For instance, they are known to learn fast from small sets of examples, while simultaneously possessing the capability of handling large data sets, and are computationally efficient to train.

One set of tree ensemble methods that have been particularly successful are gradient boosted regression trees (GBRT) \cite{Friedman2001}. This type of tree ensemble use the concept of boosting \cite{schapire1999} to construct ensembles of trees that one-by-one reduce the error of the previous trees. By only allowing for the constituent trees to grow to a certain depth, this method is able to learn complex patterns in the training data without significant overfitting to noise. In order to build such ensemble of error-correcting trees, a GBRT model starts with a base prediction $f_0$ which is just a constant value---often selected as the average of the target values in the training set. Then, for each boosting iteration $n$, the model builds a regression tree $f_n$ with splitting criteria based on the available features, aiming to group the data points into leaves, where all other data points display a similarity in their feature values, and provide a similar contribution to the total prediction error. Each leaf then calculates an output value that pushes the prediction on those data points towards a lower error. After $N$ boosting iterations, the GBRT is represented by the additive model $F_N$:
\begin{equation} \scalebox{1.0}{%
    $
    F_N(\mathbf{x}) = f_0 + \sum_{n=1}^N \xi f_n(\mathbf{X}),
    $}\label{eq_gbrt}
\end{equation}
where $\xi$ is the learning rate parameter controlling the step size in the trees' error corrections.

\subsubsection{XGBoost} 
\begin{figure}[!t]
\centering
\includegraphics[width=\textwidth]{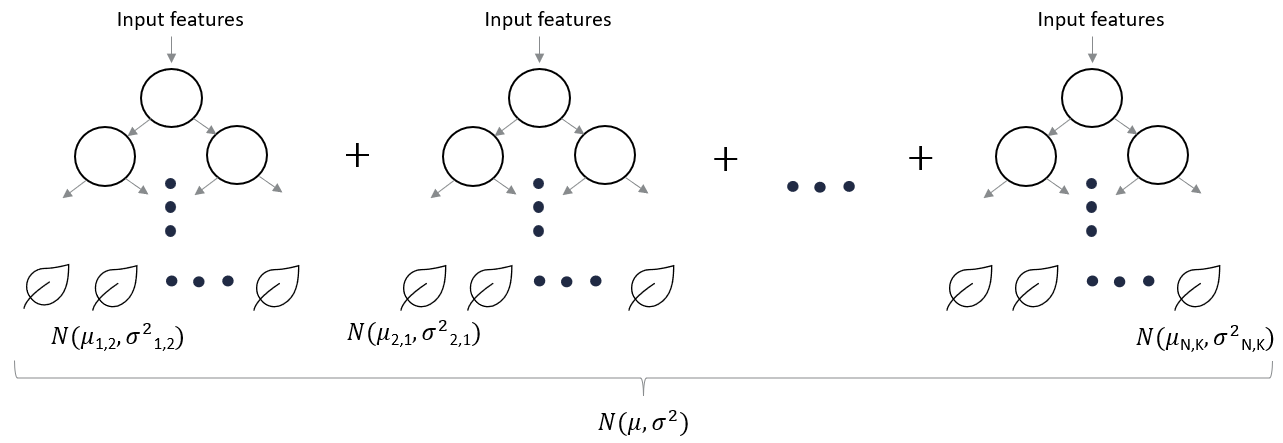}
\caption{Schematic view of the predictive distribution calculation with XGBoost tree ensemble.}
\label{fig:xgb}
\end{figure}

Extreme gradient boosting \cite{xgboost}, also known as XGBoost, is a highly effective method for building gradient boosted regression trees, which is optimized for speed and accuracy in the function fitting. XGBoost has proven itself useful many times by being the top-performing model on machine learning competitions hosted on the Kaggle\footnote{https://www.kaggle.com/} platform. An XGBoost model only produces a single output, which after building it using our training set, we may consider as the mean predictions $\mu_l$. Hence, by default there we do not obtain any variance estimates from the XGBoost model. However, \cite{nilsson2024} demonstrated a method to extract the predictive variance from a trained tree ensemble such as XGBoost and consider the output as a predictive Gaussian distribution. Since the leaf values of each tree of the ensemble are based on the sample mean of the residuals from the training data points assigned to them, one may also consider the sample variance therein. Then, similar to how the total mean prediction is calculated, the variance is considered as the sum of each tree's individual variance prediction. A schematic view of this approach is shown in Figure \ref{fig:xgb}.

\subsubsection{NGBoost} Another method developed for building gradient boosted regression trees is natural gradient boosting, or NGBoost \cite{duan2020}. A neat aspect of NGBoost is that it is specifically developed for outputting a full parameterized predictive distribution, rather than just a point estimate. In order to do so, NGBoost considers multiple outputs as the set of parameters of a pre-selected probability distribution. In our case, we again consider the energy consumption over the links as normally distributed, fully determined by its mean and variance. During the boosting iterations of the model building phase, the trees are fitted to minimize a scoring rule which compares the estimated probability distribution to the observed data. The method's name comes from the search process for the parameters that minimize the scoring rule, where NGBoost make use of a concept from information geometry called natural gradients \cite{amari1998}. As our choice of scoring rule, we use the Gaussian negative log-likelihood function, shown in Equation \ref{eq_gnll}, as for the neural networks.

In contrast to neural networks which learn smooth functions of the input features, tree ensemble models build piece-wise constant functions based on the statistics of groupings of the training data. Therefore, it is not straight-forward how to tune the multiplicative factors of the underlying physical equations for energy consumption \ref{phys_eq_energy}, as we could do using neural networks. Instead, in order to incorporate this prior information into the XGBoost and NGBoost models, we first predict the mean energy consumption using the physical equation with prior parameters $\mathbf{m}_0$, where after we fit the respective tree ensembles to correct the errors and predict the corresponding variances on the links based on the observed values $e_l$.

\subsection{Reference model}
As a baseline to compare the six machine learning models described above to, we additionally compute predictions using the linear Equation \ref{phys_eq_linreg} with parameters $\mathbf{m}_0$. We refer to this model as the reference model, or REF, and it corresponds to a standard physics-based approach to the problem. This is also the same model we use as the prior for the Bayesian linear regression model, as well as initial predictions for the gradient boosted regression trees.

\section{Experiments}
To test out the devised machine learning models and compare them to each other and the reference method, we use the data collected for all features presented in Table \ref{tab:features} also for the test set. Although accurately predicting features such as speed and velocity is both challenging and crucial to make good predictions of energy consumption in future routes, our goal is to provide a fair comparison of the different machine learning models. Therefore, to avoid poor feature prediction distort the results, we assume that the features can be predicted to the accuracy of the measured data. That being said, there will always be inherent noise from the sensors, and so the results at least to some degree indicate the models' abilities to distinguish signal from noise through the training procedure and make robust predictions from data with a similar level of noise.

It should be noted here that our main objective is accurate prediction of energy consumption and the uncertainty thereof. Estimation of the underlying physical parameters, which is an important part of some of the machine learning models, is a means to this end. This way of incorporating physical information into the data-driven methods serves two purposes: Firstly, it acts to regularize the models to help them filter out the signal from the noise in the measured data. Secondly, it makes the model somewhat more interpretable in the sense that we can observe how much the models emphasizes the different known physical effects and reason about whether it is reasonable. In order to learn trust-worthy estimates of the physical parameters, without having to handle the complexity of energy losses due to mechanical braking, or energy regained through regenerative braking, we filter out all data points where either of these forms of braking is applied. We note, however, that this is a crucial part of energy consumption prediction in practice, which we elaborate on in the future work section.

\subsection{Metrics}
After the models have been build using the training and validation routes, we set out to compare their performances on the $R$ test routes using four different metrics: Mean Absolute Percentage Error (MAPE), Maximum Absolute Percentage Error (MaxAPE), Mean Percentage Standard Deviation (MPSD) and Coverage Probability (CP).

\subsubsection{Mean Absolute Percentage Error} The Mean Absolute Percentage Error measures the accuracy of forecasting energy consumption and is calculated as 
\begin{equation} \scalebox{1.0}{%
    $
        \frac{1}{R} \sum_{r=1}^R \left| \frac{e_{\text{trip}_r} - \mu_{\text{trip}_r}}{e_{\text{trip}_r}} \right| \times 100,
    $}\label{eq_mape}
\end{equation}
with $e_{\text{trip}_r}$ and $\mu_{\text{trip}_r}$ being the true measured and the predicted energy consumption throughout route $r$ respectively.

\subsubsection{Maximum Absolute Percentage Error} The Maximum Absolute Percentage Error is used to provide a measure of the worst-case prediction of each respective model, and is calculated as
\begin{equation} \scalebox{1.0}{%
    $
        \text{max}_{r \in [R]} \left( \left| \frac{e_{\text{trip}_r} - \mu_{\text{trip}_r}}{e_{\text{trip}_r}} \right| \times 100 \right).
    $}\label{eq_maxape}
\end{equation}
\subsubsection{Mean Percentage Standard Deviation} The Mean Percentage Standard Deviation quantifies how much the models expect that the true energy consumption will deviate from their mean predictions. We report this value in percentage of the predicted mean
\begin{equation} \scalebox{1.0}{%
    $
        \frac{1}{R} \sum_{r=1}^R \frac{\sigma_{\text{trip}_r}}{\mu_{\text{trip}_r}} \times 100.
    $}\label{eq_mpsd}
\end{equation}
Ultimately, this quantity should be low, which signals that the models are able to capture the complex relationships between the features and energy consumption. However, we must make sure that the standard deviation predicted by the models also reflects the underlying uncertainty that may still exist, which leads us to to the fourth and last criteria.

\subsubsection{Coverage Probability} The Coverage Probability (CP) gives a measure of how well the models are able to capture the uncertainty and provide accurate estimates of the standard deviation. It simply calculates the fraction of the $R$ test routes for which the measured energy consumption $e_{\text{trip}_r}$ lies within $\mu_{\text{trip}_r} \pm q \sigma_{\text{trip}_r}$. A standard coverage probability to consider is $\text{CP}^{95}$, which by our Gaussian assumption means that for $95\%$ of all routes we predict the energy consumption for, the measured value should lie within the predicted mean plus/minus $1.96$ standard deviations. Hence, $\text{CP}^{95}$ provides an estimate of how well-calibrated our models are, and a small MPSD is only valuable as long as the $\text{CP}^{95}$ is close to $95\%$.

\section{Results and discussion}\label{sec:discussion}
Although each truck has its own set of machine learning models, trained only on the data is has generated itself, we present here the results averaged over all routes from the four trucks together. This is to assess the effectiveness of the different machine learning approaches in general. In \ref{appendix:additional_results}, we also present the results for each truck individually, so as to gain insight into how stable the models' performances are. To mitigate sample bias, the results are also averaged over 10 different runs of randomly selecting routes for training, validation and testing. Table \ref{tab:results_average} shows how the different models perform on the four criteria described above.

\begin{table}[t]
\setlength{\tabcolsep}{0.5\tabcolsep}
\centering
\begin{tabular}{l c c c c}
\toprule
\textbf{Model} & \textbf{MAPE (\%)} & \textbf{MaxAPE (\%)} & \textbf{MPSD (\%)} & $\textbf{CP}^{95}$\\
\midrule
\textbf{REF} & $12.19 \pm 1.56$ & $19.62 \pm 3.05$ & $17.53 \pm 2.92$ & $0.96 \pm 0.07$ \\
\textbf{MLR} & $4.93 \pm 1.38$ & $14.18 \pm 6.33$ & $8.16 \pm 3.12$ & $0.92 \pm 0.11$ \\
\textbf{BMLR} & $4.81 \pm 1.06$ & $12.53 \pm 4.74$ & $7.49 \pm 2.23$ & $0.91 \pm 0.12$ \\
\textbf{NN} & $4.64 \pm 1.52$ & $11.75 \pm 4.37$ & $7.66 \pm 2.04$ & $0.92 \pm 0.12$ \\
\textbf{BNN} & $4.53 \pm 1.25$ & $12.09 \pm 3.52$ & $6.17 \pm 2.38$ & $0.89 \pm 0.12$ \\
\textbf{XGBoost} & $3.38 \pm 0.90$ & $9.95 \pm 4.39$ & $5.35 \pm 1.82$ & $0.90 \pm 0.14$ \\
\textbf{NGBoost} & $3.42 \pm 0.90$ & $10.29 \pm 4.71$ & $6.91 \pm 2.61$ & $0.90 \pm 0.13$ \\
\bottomrule
\end{tabular}
\caption{Performance of physics-aware machine learning models for energy consumption prediction on previously unseen routes from four different trucks.}
\label{tab:results_average}
\end{table}

In Table \ref{tab:params_average}, we present the average percent difference between the parameters learned by the linear and neural network based models, as compared to the predetermined reference values of the trucks. In \ref{appendix:additional_results}, this information is presented for each truck individually. As discussed when introducing the models, the linear ones learn the average parameter values under the operating conditions of the trucks in the training routes, while the neural networks have the ability to learn dynamic mappings of the input features to instantaneous parameter values. Therefore, when reporting the results here for the percent difference of the parameter values learned by the neural networks, we first average the instantaneous parameter values that the networks predicted on each of the segments in the test routes. 

\begin{table}[t]
\centering
\begin{tabular}{l c c c}
\toprule
\textbf{Model} & \textbf{$\frac{1}{\eta}$ (\%)} & \textbf{$\frac{C_r}{\eta}$ (\%)} & \textbf{$\frac{C_d}{\eta}$ (\%)} \\
\midrule
\textbf{MLR} & 15.85 & 280.55 & 100.02 \\

\textbf{BMLR} & 4.93 & 33.62 & 23.86 \\

\textbf{NN} & 3.12 & 46.64 & 24.01 \\

\textbf{BNN} & 2.79 & 52.80 & 23.80 \\
\bottomrule
\end{tabular}
\caption{Percent difference of the learned physical parameters compared to reference values}
\label{tab:params_average}
\end{table}

When evaluating these results, one must consider that the learned quantities from any of the models may not correspond directly to the true underlying physical values that they represent. For instance, if there is a bias or systematic error in one or more of the sensors measuring the features that are fed to the machine learning methods, this will also be picked up by the models and accounted for in the learned factors. That being said, the results indicate a significant bias in the predetermined reference values. For each of the machine learning models, and all four of the trucks, the learned parameter values related to rolling resistance and air drag differ significantly from the reference values. The values learned by the standard linear regression model display particularly large differences here. However, there are multiple reasons to disbelieve the results for this model. Firstly, the difference is also large when it comes to the inverse powertrain efficiency (over $10 \%$) which is not true for the other models. This is a parameter that is relatively well-known and easy to determine, and so it is likely that the linear regression model has erroneously under-predicted this parameter. Secondly, looking at the percent difference for the air drag coefficient over efficiency, we observe that the linear regression model predicts it very close to zero, and for two of the trucks even slightly negative. This does not make any sense physically, and agrees with what was observed in \cite{DeCauwer2027}, as we discussed in the related work section, where linear regression tended to set the air drag coefficient to suspiciously low values. Lastly, the parameter values found by the other machine learning methods appear to agree quite well with each other here, which further strengthens our beliefs that the ones determined by the linear regression model are not to be trusted. 

Considering the implausible parameter values of the standard linear regression model, it is interesting to notice that this model manages to perform fairly well on our different performance metrics for energy consumption prediction presented in Table \ref{tab:results_average}. The model seems to capture a general trend that the other machine learning methods hint at, which is that the reference value for rolling resistance is often set too low. Although not being able to find parameters that make sense physically, it appears that this model is still able to reduce much of the bias present in the reference model. Comparing the results of the standard linear regression model with those of the Bayesian linear regression, first in Table \ref{tab:results_average}, the Bayesian version seems to have a slight edge when it comes to predicting the expected energy consumption. In addition, the results in Table \ref{tab:params_average} indicate that the Bayesian linear regression model is able to learn much more realistic parameter values, which strengthens its reliability. We believe this to be a key reason why the difference in performance is particularly large when it comes to the Maximum Absolute Percentage Error.

Moving to the neural network models, with the increased flexibility of predicting non-stationary parameter values and including more of the features from Table \ref{tab:features}, there seem to be little or no gain in terms of accuracy in energy consumption prediction and predictive uncertainty. As in the case of the linear models, the Bayesian version of the physics-aware neural networks appear to perform slightly better in terms of MAPE. However, the non-Bayesian equivalent displays a smaller MaxAPE, although the difference is minimal. Table \ref{tab:params_average} offers an explanation for the tiny differences here, as both neural network models appear to have learned similar parameters of the physical model. These values are also relatively close to the ones discovered by the BMLR model, and hence physically feasible. When it comes to MPSD and $\text{CP}^{95}$, all of the linear and neural networks based models perform similar, and there is an apparent tendency of the models to under-estimate the variance, as signified by the relatively low $\text{CP}^{95}$ values. The same observations of somewhat poorly calibrated variance estimates are present for the XGBoost and NGBoost models. However, the limited number of routes makes it difficult to draw strong conclusions about the difference in model performances from the $\text{CP}^{95}$ metric, and the fact that all models present a value close to $0.95$ on average is a good indication that the overall approach is reasonable. An interesting observation to make is that all of the machine learning models report $\text{CP}^{95}$ values that are similar and smaller than 0.95 on average. This may suggest that their variance estimations are good, but that the Gaussian assumption that the method is based on is incorrect. Overall, we observe that the two gradient boosted regression tree methods perform similar to each other, and somewhat better than the other models in terms of MAPE, MaxAPE and MPSD, with a similar coverage probability.

\section{Conclusions}
To conclude, we have proposed a set of physics-aware machine learning methods with the aim to unify physics-based and data-driven approaches to vehicle energy consumption prediction, by incorporating a physical model into different function-fitting processes. Our study demonstrate an advantage of this approach as compared with standard machine learning models. The fact that the gradient boosted regression trees outperform all other models in this study may not come as a surprise to the machine learning community. Such methods often perform well on problems with structure data like we have in this case. Interestingly, these models also performed on par with the neural network based and linear models when it comes to uncertainty estimation. This is noteworthy, as these models are relatively underexplored for this matter, in particular XGBoost.

One possible way to obtain more accurate and reliable results, especially when it comes to the variance estimation and coverage probability, would be to run the methods on a larger dataset. To improve the accuracy in variance estimation one may study the covariance between links of different lengths of separation in more detail, rather than relying on the average correlation between them like we do through Equation \ref{eq_var_cov_trip}, or consider a different family of distributions than the Gaussian assumption we made in this work. Another step toward the same goal would be to consider full isotonic regression for variance calibration, as proposed by \cite{kuleshov2018}, rather than simply learning a multiplicative factor. We leave these potential improvements for future work.

Another area where future work is needed is to study this problem with inputs that are not known before-hand. The errors inherent with feature predictions pose a challenge for the models when it comes to predicting energy consumption on future routes, and it is unclear for this work how well the respective models would perform in this setting. Furthermore, for increased applicability, future work would also need to include energy losses and recoveries from mechanical and regenerative braking. When extending this work to these more difficult scenarios with higher uncertainty, our results may be considered as a target performance level to aim for.

\appendix

\section{Table of notation}
To facilitate readability and provide a quick reference for the mathematical formulations discussed throughout this work, we present a comprehensive summary of the notation used. Table \ref{tab:notation} outlines the key symbols and variables, categorized into three main groups: physical parameters governing vehicle dynamics, statistical and machine learning variables associated with the predictive models, and the metrics used for performance evaluation

To facilitate readability and provide a quick reference for the mathematical formulations discussed throughout this work, we present a comprehensive summary of the notation used. Table \ref{tab:notation} outlines the key symbols and variables, categorized into three main groups: physical parameters governing vehicle dynamics, statistical and machine learning variables associated with the predictive models, and the metrics used for performance evaluation.

\renewcommand{\arraystretch}{1.2}
\begin{longtable}{ll}
\caption{Table of Notation} \label{tab:notation} \\
\hline
\textbf{Symbol} & \textbf{Description} \\
\hline
\endfirsthead

\multicolumn{2}{c}{{\bfseries \tablename\ \thetable{} -- continued from previous page}} \\
\hline
\textbf{Symbol} & \textbf{Description} \\
\hline
\endhead

\hline 
\multicolumn{2}{r}{{Continued on next page}} \\
\endfoot

\hline
\endlastfoot

\multicolumn{2}{l}{\textbf{Physical Parameters and Vehicle Dynamics}} \\
$f$ & Instantaneous wheel traction force \\
$m$ & Vehicle mass \\
$a$ & Vehicle acceleration \\
$g$ & Gravitational acceleration \\
$\theta$ & Road slope angle / road inclination \\
$C_{r}$ & Rolling resistance coefficient \\
$C_{d}$ & Aerodynamic drag coefficient \\
$A$ & Frontal area of the vehicle \\
$\rho$ & Air density \\
$v$ & Vehicle velocity / speed relative to wind \\
$v_i, v_f$ & Initial and final velocity over a short time step \\
$\eta$ & Powertrain efficiency \\
$d$ & Distance driven over a short time step \\
$e$ & Expected energy consumption over distance $d$ \\
\multicolumn{2}{c}{} \\ 

\multicolumn{2}{l}{\textbf{Statistical and Machine Learning Variables}} \\
$e_l$ & Measured energy consumption on link $l$ \\
$L$ & Total number of segments/links in a trip \\
$\mu_l, \sigma_l^2$ & Predicted mean and variance of energy consumption on link $l$ \\
$\mu_{trip}, \sigma_{trip}^2$ & Predicted mean and variance of energy consumption for a full trip \\
$c_{cov}$ & Covariance factor between links \\
$c_l$ & Multiplicative calibration factor for variance via isotonic regression \\
$D$ & Number of linear features \\
$w$ & Unknown coefficients vector in linear regression \\
$X, t$ & Observation matrix and target variables vector \\
$\beta$ & Precision of the Gaussian noise \\
$m_0, S_0$ & Prior mean vector and covariance matrix \\
$m_L, S_L$ & Posterior mean vector and covariance matrix \\
$\omega$ & Neural network weights \\
$\sigma_{\omega_0}^2, \sigma_{\omega_L}^2$ & Prior and posterior weight variance in BNNs \\
$()^{\mathrm{ref}}$ & Reference quantity obtained from prior vehicle testing \\
$()^{\mathrm{ANN}}$ & Correction term predicted by artificial neural network \\
$\alpha$ & Multiplicative relative weight factor for the KL divergence loss term \\
$\lambda$ & L2 regularization parameter (weight decay) \\
$\Omega$ & Total number of weights in the neural network \\
$f_0$ & Base prediction in gradient boosting \\
$f_n, F_N$ & Regression tree at iteration $n$ and the final additive model \\
$\xi$ & Learning rate parameter for gradient boosting \\
$\mathcal{L}_{GNLL}$ & Gaussian negative log-likelihood loss function \\
\multicolumn{2}{c}{} \\ 

\multicolumn{2}{l}{\textbf{Evaluation Metrics}} \\
$R$ & Total number of test routes \\
$e_{trip_r}$ & True measured energy consumption for route $r$ \\
$\mu_{trip_r}$ & Predicted mean energy consumption for route $r$ \\
$\sigma_{trip_r}$ & Predicted standard deviation of energy consumption for route $r$ \\
$q$ & Coverage probability multiplier (e.g., 1.96 for 95\%) \\

\end{longtable}

\section{Details on Bayesian multivariate linear regression}\label{appendix:bayesian_details}
Here, we outline how to obtain to estimate $p(\mathbf{w}|\mathbf{t})$ using Bayesian multiple linear regression. First, let $\mathbf{X} = \big[\mathbf{x}_1, \mathbf{x}_1,..., \mathbf{x}_L\big]$ denote the observation matrix, where each $\mathbf{x}_l$ represents a vector of observed features:
\begin{equation} \scalebox{1.0}{%
    $
    \mathbf{x}_l = \Big[(mad + mgd \sin \theta)_l, (mg d \cos \theta)_l, \left(\rho d \left(v_i^2 + v_f^2\right)\right)_l\Big]^T.
    $}\label{eq_features}
\end{equation}
Now, we assume that the target variables $t$ are given by
\begin{equation} \scalebox{1.0}{%
    $
    e = \frac{1}{\eta}(mad + mgd \sin \theta) + \frac{C_r}{\eta}mg d \cos \theta + \frac{C_d A}{4 \eta} \rho d \left(v_i^2 + v_f^2\right)
    $}\label{phys_eq_linreg_appendix},
\end{equation}
 with additive Gaussian noise. After placing the coefficients to be determined in a weight vector
 \begin{equation} \scalebox{1.0}{%
    $
    \mathbf{w} = \Big[\frac{1}{\eta}, \frac{C_r}{\eta}, \frac{C_d A}{4 \eta} \Big]^T
    $}\label{weight_vector},
\end{equation}
 we therefore have:
\begin{equation} \scalebox{1.0}{%
    $
    t = \mathbf{w}^T \mathbf{x} + \epsilon
    $}\label{eq_targets},
\end{equation}
where $\epsilon$ is a zero mean Gaussian random variable with precision $\beta$. The probability of a target value $t$ given a set of features $\mathbf{x}$ and coefficients $\mathbf{w}$ is then given by the following Gaussian distribution:
\begin{equation} \scalebox{1.0}{%
    $
    p(t|\mathbf{x}, \mathbf{w}, \beta) = \mathcal{N}(t|\mathbf{w}^T \mathbf{x}, \beta^{-1})
    $}\label{eq_targets_prob}.
\end{equation}
If we further assume that all data points are drawn independently from Equation \ref{eq_targets_prob} (which is not exactly true, as we discuss later regarding uncertainty estimation), we get the following expression for the likelihood of the series of energy measurements $\mathbf{t} = [t_1, t_2,...,t_L]^T$:
\begin{equation} \scalebox{1.0}{%
    $
    p(\mathbf{t}|\mathbf{x}, \mathbf{w}, \beta) = \prod_{n=1}^L\mathcal{N}(t_l|\mathbf{w}^T \mathbf{x}_l, \beta^{-1})
    $}\label{eq_series_targets}.
\end{equation}
Next, we model the prior $p(\mathbf{w})$ as a Gaussian distribution with mean vector $\mathbf{m}_0$ and covariance matrix $\mathbf{S}_0$:
\begin{equation} \scalebox{1.0}{%
    $
    p(\mathbf{w}) = \mathcal{N}(\mathbf{w}|\mathbf{m}_0,\mathbf{S}_0).
    $}\label{eq_prior}
\end{equation}
The set of assumptions we have made lead to nice and simple calculations of the posterior distribution in the following form:
\begin{equation} \scalebox{1.0}{%
    $
    p(\mathbf{w}|\mathbf{t}) = \mathcal{N}(\mathbf{w}|\mathbf{m}_L,\mathbf{S}_L),
    $}\label{eq_posterior}
\end{equation}
with 
\begin{equation} \scalebox{1.0}{%
    $
    \mathbf{m}_L = \mathbf{S}_L (\mathbf{S}_0^{-1} \mathbf{m}_0 + \beta \mathbf{X}^T \mathbf{t} ),
    \quad
    \mathbf{S}_L = \mathbf{S}_0^{-1} + \beta \mathbf{X}^T\mathbf{X}.
    $}\label{eq_m_N_S_N}
\end{equation}
Recall that it is the posterior distribution over coefficients that we are after in order to make future prediction of energy consumption and the associated predictive variance.

This Bayesian version of multi-dimensional linear regression has a couple of nice properties that makes it interesting to study in this context. Since it allows for specifying a prior distribution, it is possible to incorporate knowledge about the problem that one might have beforehand. In our case, we use information from the specific truck models and tires to set the prior mean values $\mathbf{m}_0$ based on measurement from controlled experiments. These values may not correspond to the exact values under the current status of the truck and driving conditions, but provide a reasonable initial guess. Such informative priors may lead to faster learning from less data, and better resilience to noise in the data as compared to standard multiple linear regression.

In addition to $\mathbf{m}_0$, the method also requires that $\mathbf{S}_0$ and $\beta$ be specified beforehand. Here, we choose $\beta = 1 / \text{Var}[\mathbf{t}]$, the reciprocal of the variance in measured energy consumption, which gives an overestimation of the noise. For $S_0$, we assume it to be diagonal and pick its diagonal elements by running the learning algorithm on a subset of the training data, and evaluate on routes from the validation set. We then pick values that seem to strike a good balance between learning fairly quickly from new data, while not overfitting to noise to provide physically implausible values.

\section{Feature importance analysis}
To better understand the decision-making process of the proposed data-driven models, we employ SHapley Additive exPlanations (SHAP) \cite{shap} to evaluate feature importance. Due to their complex, non-linear architectures, advanced machine learning models like gradient boosted regression trees often act as "black boxes," making it difficult to interpret how individual inputs influence the final prediction. SHAP provides a rigorous framework to resolve this by quantifying the exact contribution of each feature. We apply it to the XGBoost models for each truck here, as they were the models that performed the best on average and give the most insight into which features are most relevant for accurate energy prediction. However, a SHAP analysis can in principle be performed on all of the models.

SHAP is a game-theoretic approach designed to explain the output of machine learning models. It is grounded in Shapley values from coalitional game theory, which were originally formulated to fairly distribute a total payout among players in a cooperative game based on their individual contributions. In the context of machine learning, the game is the prediction task for a single instance, the total payout is the difference between the actual prediction and the baseline (expected) prediction, and the players are the input features. SHAP calculates the importance of a feature by evaluating model predictions across all possible combinations (or coalitions) of features. Specifically, it computes the marginal contribution of a feature by observing how the prediction changes when that feature is added to a model that only contains a subset of the other features. The final SHAP value for a specific feature is the weighted average of all its marginal contributions across all possible feature permutations. A higher absolute SHAP value indicates that the feature has a larger impact on the model's energy consumption prediction for that specific road segment.

\begin{figure}
    \centering
    \subfloat[Truck 1]{\includegraphics[width=0.48\textwidth]{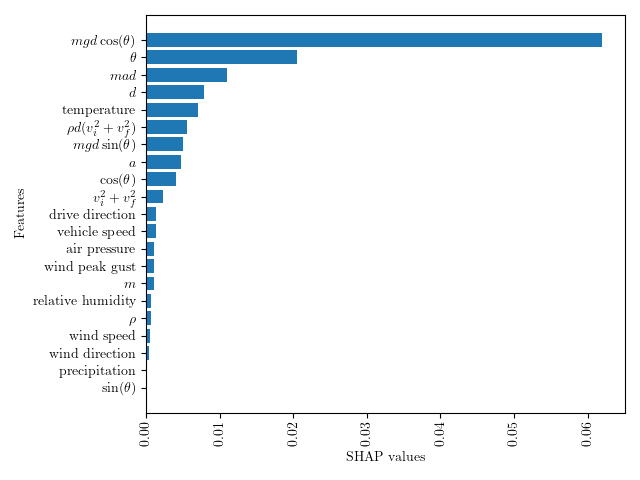}} 
    \subfloat[Truck 2]{\includegraphics[width=0.48\textwidth]{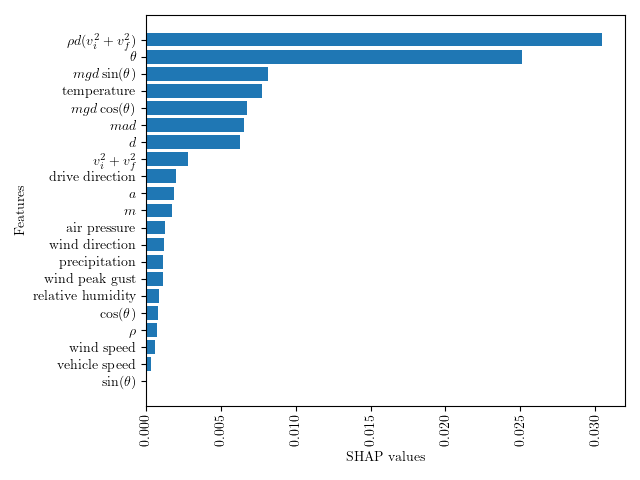}} 
    
    \subfloat[Truck 3]{\includegraphics[width=0.48\textwidth]{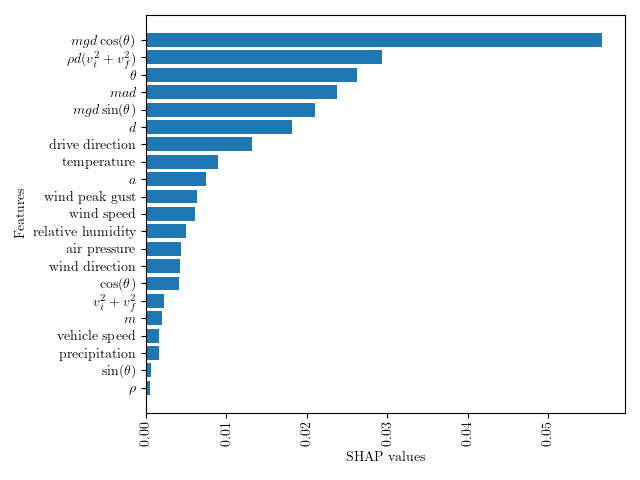}}
    \subfloat[Truck 4]{\includegraphics[width=0.48\textwidth]{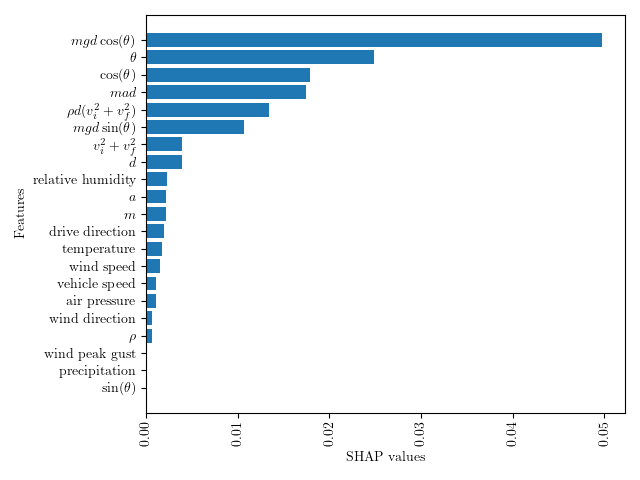}}
    \caption{SHAP values for the input features of the XGBoost models for each respective truck.}
    \label{fig:shap_values}
\end{figure}

Figure \ref{fig:shap_values} displays the mean absolute SHAP values for the input features utilized by the XGBoost models across the four different trucks. Evaluating these global feature importances yields several key insights into how the physics-aware machine learning models operate.

Firstly, we observe a dominance of the physical multipliers. Across all four trucks, the composite features derived directly from the physical relationships in \ref{phys_eq_energy}, i.e. the rolling resistance term ($mgd \cos(\theta)$), the aerodynamic drag term ($\rho d(v_i^2 + v_f^2)$), $mad$ and $mgd \sin(\theta)$ play a significant role for the predictions. It is not surprising that these terms, which act as the linear multipliers in the baseline physical equations, are at the top of the feature importance hierarchy. As demonstrated by our experiments with the linear machine learning models, the predetermined physical reference values for these trucks contained significant biases. Consequently, the gradient boosted models heavily rely on these fundamental physical terms to correct the baseline offsets and establish much more accurate mean predictions.

Beyond the primary physical terms, many of environmental and weather-related variables also display notable SHAP values. The inclusion of these additional features is likely to account for the improvement in prediction compared with the linear models, as they complement the linear physical model by dynamically scaling the predictions based on real-world conditions. For example, temperature inherently affects tire pressure, road friction, and air density in ways that fluctuate continuously. By assigning importance to these variables, the machine learning models successfully account for complex, non-stationary dynamic factors that are entirely absent from traditional physics-based formulas.

Additionally, we observe that some of the measured physical quantities, such as the road inclination $\theta$ and distance $d$ tend to have a significant effect on the final predictions. One explanation for this might again be that the physical model of Equation \ref{phys_eq_energy} represents a simplification and does not account for all sources of energy consumption. Hence, the XGBoost models might leverage these variables to discover some dependencies that are not explicitly covered by the model used. Furthermore, the devices used to measure these quantities might have an inherent bias, providing errors when inserted into the physical formula. For instance, road inclination is tricky to measure accurately, and so this provides another explanation for the large SHAP values assigned to $\theta$.

In summary, the SHAP analysis indicates that the XGBoost models successfully merge theoretical vehicle dynamics with data-driven flexibility. They prioritize the governing physical laws of motion to anchor their predictions, while effectively utilizing environmental data to capture the detailed, non-linear realities of field operations.

\section{Additional results}\label{appendix:additional_results}
Tables \ref{tab:results} and \ref{tab:params} show the results from the experiments described in the main paper for each truck individually. The Tables show that the models' performances are generally stable, with similar results for each truck.

\begin{table}[ht]
\setlength{\tabcolsep}{0.5\tabcolsep}
\centering
\begin{tabular}{l c c c c}
\toprule
\textbf{Model} & \textbf{MAPE (\%)} & \textbf{MaxAPE (\%)} & \textbf{MPSD (\%)} & $\textbf{CP}^{95}$\\
\midrule
\textbf{REF} & & & & \\
\quad Truck 1  &  $4.03 \pm 1.33$  &  $10.53 \pm 4.55$  & $5.97 \pm 2.43$ &  $0.80 \pm 0.30$ \\
\quad Truck 2  &  $5.36 \pm 1.72$  &  $11.32 \pm 2.71$  & $9.09 \pm 2.29$ &  $0.97 \pm 0.09$ \\
\quad Truck 3  &  $13.01 \pm 2.08$  &  $20.43 \pm 3.42$  & $18.18 \pm 3.16$ &  $0.99 \pm 0.04$ \\
\quad Truck 4  &  $17.78 \pm 1.29$  &  $26.21 \pm 2.43$  & $25.39 \pm 2.32$ &  $1.00 \pm 0.00$ \\
\textbf{MLR} & & & & \\
\quad Truck 1  &  $3.53 \pm 1.38$  &  $9.71 \pm 7.95$  & $6.00 \pm 2.71$ &  $0.92 \pm 0.14$ \\
\quad Truck 2  &  $5.14 \pm 1.22$  &  $13.20 \pm 1.85$  & $8.04 \pm 3.61$ &  $0.93 \pm 0.10$ \\
\quad Truck 3  &  $7.07 \pm 2.82$  &  $22.81 \pm 16.18$  & $12.35 \pm 4.59$ &  $0.94 \pm 0.06$ \\
\quad Truck 4  &  $4.16 \pm 0.65$  &  $11.42 \pm 2.13$  & $7.41 \pm 2.88$ &  $0.91 \pm 0.13$ \\
\textbf{BMLR} & & & & \\
\quad Truck 1  &  $3.71 \pm 0.97$  &  $9.39 \pm 4.60$  & $5.76 \pm 2.44$ &  $0.87 \pm 0.17$ \\
\quad Truck 2  &  $3.65 \pm 0.99$  &  $8.12 \pm 2.52$  & $5.14 \pm 3.20$ &  $0.84 \pm 0.23$ \\
\quad Truck 3  &  $7.04 \pm 1.75$  &  $19.02 \pm 10.70$  & $11.84 \pm 3.23$ &  $0.93 \pm 0.09$ \\
\quad Truck 4  &  $4.48 \pm 0.74$  &  $11.98 \pm 2.42$  & $7.32 \pm 2.31$ &  $0.94 \pm 0.06$ \\
\textbf{NN} & & & & \\
\quad Truck 1  &  $4.84 \pm 2.23$  &  $11.35 \pm 5.43$  & $6.19 \pm 2.52$ &  $0.88 \pm 0.11$ \\
\quad Truck 2  &  $4.55 \pm 2.36$  &  $10.66 \pm 6.92$  & $8.45 \pm 3.11$ &  $0.96 \pm 0.10$ \\
\quad Truck 3  &  $7.53 \pm 1.63$  &  $19.14 \pm 6.38$  & $12.70 \pm 4.86$ &  $0.93 \pm 0.12$ \\
\quad Truck 4  &  $2.97 \pm 0.82$  &  $8.22 \pm 1.72$  & $4.55 \pm 1.77$ &  $0.91 \pm 0.13$ \\
\textbf{BNN} & & & & \\
\quad Truck 1  &  $4.47 \pm 1.28$  &  $9.87 \pm 4.09$  & $5.84 \pm 3.19$ &  $0.85 \pm 0.17$ \\
\quad Truck 2  &  $3.70 \pm 1.43$  &  $8.29 \pm 2.78$  & $6.35 \pm 3.74$ &  $0.90 \pm 0.14$ \\
\quad Truck 3  &  $7.49 \pm 2.01$  &  $21.32 \pm 6.90$  & $10.56 \pm 2.51$ &  $0.93 \pm 0.08$ \\
\quad Truck 4  &  $3.26 \pm 0.74$  &  $9.40 \pm 1.74$  & $4.52 \pm 1.19$ &  $0.89 \pm 0.11$ \\
\textbf{XGBoost} & & & & \\
\quad Truck 1  &  $3.59 \pm 0.93$  &  $8.22 \pm 3.68$  & $4.73 \pm 2.49$ &  $0.82 \pm 0.31$ \\
\quad Truck 2  &  $2.46 \pm 0.92$  &  $6.84 \pm 2.59$  & $4.77 \pm 2.94$ &  $0.87 \pm 0.14$ \\
\quad Truck 3  &  $4.81 \pm 1.42$  &  $15.07 \pm 7.85$  & $7.91 \pm 1.38$ &  $0.91 \pm 0.10$ \\
\quad Truck 4  &  $2.91 \pm 0.58$  &  $9.08 \pm 3.49$  & $5.36 \pm 1.52$ &  $0.94 \pm 0.09$ \\
\textbf{NGBoost} & & & & \\
\quad Truck 1  &  $3.34 \pm 1.07$  &  $8.55 \pm 4.25$  & $4.44 \pm 1.52$ &  $0.83 \pm 0.21$ \\
\quad Truck 2  &  $2.42 \pm 0.95$  &  $6.42 \pm 2.06$  & $4.53 \pm 2.93$ &  $0.87 \pm 0.14$ \\
\quad Truck 3  &  $4.49 \pm 1.32$  &  $15.79 \pm 10.03$  & $7.62 \pm 3.95$ &  $0.86 \pm 0.17$ \\
\quad Truck 4  &  $3.28 \pm 0.57$  &  $9.54 \pm 3.04$  & $6.42 \pm 1.23$ &  $0.97 \pm 0.07$ \\
\bottomrule
\end{tabular}
\caption{Performance of physics-aware machine learning models for energy consumption prediction on previously unseen routes. The results presented are the averages plus/minus one standard deviation over 10 runs of different random splitting of routes for training, validation and training.}
\label{tab:results}
\end{table}

\begin{table}[ht]
\centering
\begin{tabular}{l c c c}
\toprule
\textbf{Model} & \textbf{$\frac{1}{\eta}$} (\%) & \textbf{$\frac{C_r}{\eta}$} (\%) & \textbf{$\frac{C_d}{\eta}$} (\%) \\
\midrule
\textbf{MLR} & & & \\
\quad Truck 1 & 10.91 & 257.84 & -99.99 \\
\quad Truck 2 & 16.30 & 188.55 & -100.07 \\
\quad Truck 3 & 25.89 & 357.65 & -100.01 \\
\quad Truck 4 & 10.31 & 318.17 & -100.00 \\

\textbf{BMLR} & & & \\
\quad Truck 1 & 4.24 & 24.81 & -3.30 \\
\quad Truck 2 & 2.93 & -7.37 & -45.47 \\
\quad Truck 3 & 9.07 & 50.20 & 9.27 \\
\quad Truck 4 & 3.48 & 52.11 & -37.40 \\

\textbf{NN} & & & \\
\quad Truck 1 & -4.03 & 5.18 & -7.47 \\
\quad Truck 2 & 0.55 & -10.70 & -40.44 \\
\quad Truck 3 & 7.86 & 92.67 & -0.27 \\
\quad Truck 4 & -0.02 & 78.01 & -47.87 \\

\textbf{BNN} & & & \\
\quad Truck 1 & -0.96 & 18.09 & -3.76 \\
\quad Truck 2 & -0.08 & -20.27 & -41.77 \\
\quad Truck 3 &  9.33 & 95.33 & 0.20 \\
\quad Truck 4 & -0.79 & 77.52 & -49.45 \\
\bottomrule
\end{tabular}
\caption{Percent difference of the learned average values of parameters in the physical model for energy consumption prediction, by different machine learning models, as compared to experimentally determined reference values.}
\label{tab:params}
\end{table}

In Tables \ref{tab:results_standard_average} and \ref{tab:results_standard}, we present the results from the same experimental setup as that described in the main paper, but with standard machine learning architectures. The Tables show that these models, which have not been design to incorporate physical information about vehicle dynamics into their architecture, perform significantly worse than their corresponding physics-aware versions proposed in this work. This is particularly true for the neural network models, where the standard versions are shown to be less stable, as signified by larger variance over the 10 different runs as well as MaxAPE values. One apparent reason for this relatively poor performance is that these models overfit to the routes they are trained and validated on. During the hyper-parameter tuning of the models, we noted that the standard ones performed better with larger regularization parameters which supports this explanation. As shown by \cite{zhu2024}, standard versions of both neural networks and gradient boosted regression trees can perform well for energy consumption prediction. However, their study relied on a huge set of routes. In this study, where the aim is to learn functions of energy consumption for individual trucks from relatively few routes, the models benefit greatly from being designed around the underlying physics governing the energy losses.

\begin{table}[ht]
\setlength{\tabcolsep}{0.5\tabcolsep}
\centering
\begin{tabular}{l c c c c}
\toprule
\textbf{Model} & \textbf{MAPE (\%)} & \textbf{MaxAPE (\%)} & \textbf{MPSD (\%)} & $\textbf{CP}^{95}$\\
\midrule
\textbf{NN} & $10.52 \pm 3.54$ & $25.15 \pm 11.91$ & $13.56 \pm 5.72$ & $0.86 \pm 0.20$ \\
\textbf{BNN} & $10.47 \pm 3.67$ & $25.35 \pm 12.28$ & $13.42 \pm 5.90$ & $0.84 \pm 0.20$ \\
\textbf{XGBoost} & $6.02 \pm 2.83$ & $14.75 \pm 8.87$ & $9.09 \pm 6.78$ & $0.91 \pm 0.12$ \\
\textbf{NGBoost} & $4.19 \pm 1.53$ & $12.85 \pm 7.56$ & $6.45 \pm 3.58$ & $0.89 \pm 0.11$ \\
\bottomrule
\end{tabular}
\caption{Performance of standard machine learning models, without physical information incorporated into their design, for energy consumption prediction on previously unseen routes from four different trucks. The results presented are the averages plus/minus one standard deviation over 10 runs of varying random splitting of routes for training, validation and training.}
\label{tab:results_standard_average}
\end{table}

\begin{table}[ht]
\setlength{\tabcolsep}{0.5\tabcolsep}
\centering
\begin{tabular}{l c c c c}
\toprule
\textbf{Model} & \textbf{MAPE (\%)} & \textbf{MaxAPE (\%)} & \textbf{MPSD (\%)} & $\textbf{CP}^{95}$\\
\midrule
\textbf{NN} & & & & \\
\quad Truck 1 & $8.58 \pm 3.76$ & $20.08 \pm 16.17$ & $12.53 \pm 6.28$ & $0.80 \pm 0.23$ \\
\quad Truck 2 & $26.12 \pm 7.21$ & $64.05 \pm 24.37$ & $32.96 \pm 9.01$ & $0.94 \pm 0.07$ \\
\quad Truck 3 & $9.12 \pm 4.70$ & $23.80 \pm 16.78$ & $11.47 \pm 6.33$ & $0.86 \pm 0.20$ \\
\quad Truck 4 & $5.22 \pm 1.20$ & $10.80 \pm 2.13$ & $7.90 \pm 2.35$ & $0.84 \pm 0.25$ \\

\textbf{BNN} & & & & \\
\quad Truck 1 & $8.52 \pm 3.73$ & $20.28 \pm 17.12$ & $12.11 \pm 5.74$ & $0.82 \pm 0.21$ \\
\quad Truck 2 & $26.22 \pm 7.03$ & $63.88 \pm 24.66$ & $33.65 \pm 9.86$ & $0.94 \pm 0.07$ \\
\quad Truck 3 & $9.21 \pm 5.07$ & $24.63 \pm 17.34$ & $11.03 \pm 8.14$ & $0.88 \pm 0.14$ \\
\quad Truck 4 & $5.01 \pm 1.40$ & $10.80 \pm 2.20$ & $6.67 \pm 2.21$ & $0.78 \pm 0.28$ \\

\textbf{XGBoost} & & & & \\
\quad Truck 1 & $7.10 \pm 5.35$ & $24.47 \pm 25.34$ & $9.34 \pm 10.09$ & $0.82 \pm 0.20$ \\
\quad Truck 2 & $3.99 \pm 2.83$ & $11.44 \pm 10.44$ & $7.50 \pm 11.24$ & $0.86 \pm 0.20$ \\
\quad Truck 3 & $12.20 \pm 5.28$ & $20.14 \pm 6.64$ & $18.76 \pm 8.19$ & $0.96 \pm 0.08$ \\
\quad Truck 4 & $3.02 \pm 0.51$ & $9.52 \pm 3.25$ & $4.32 \pm 1.77$ & $0.93 \pm 0.07$ \\

\textbf{NGBoost} & & & & \\
\quad Truck 1 & $5.54 \pm 4.26$ & $21.05 \pm 24.48$ & $7.45 \pm 7.98$ & $0.80 \pm 0.17$ \\
\quad Truck 2 & $2.93 \pm 1.16$ & $9.80 \pm 5.07$ & $4.68 \pm 3.07$ & $0.87 \pm 0.13$ \\
\quad Truck 3 & $5.89 \pm 1.63$ & $15.63 \pm 5.49$ & $9.11 \pm 3.37$ & $0.92 \pm 0.07$ \\
\quad Truck 4 & $3.28 \pm 0.62$ & $9.54 \pm 3.46$ & $5.49 \pm 1.83$ & $0.92 \pm 0.11$ \\

\bottomrule
\end{tabular}
\caption{Performance of standard machine learning models, without physical information incorporated into their design, for energy consumption prediction on previously unseen routes. The results presented are the averages plus/minus one standard deviation over 10 runs of different random splitting of routes for training, validation and training.}
\label{tab:results_standard}
\end{table}

\section{Truck and route details}\label{appendix:data_details}
\begin{figure*}[ht]
\centering
\includegraphics[width=\textwidth]{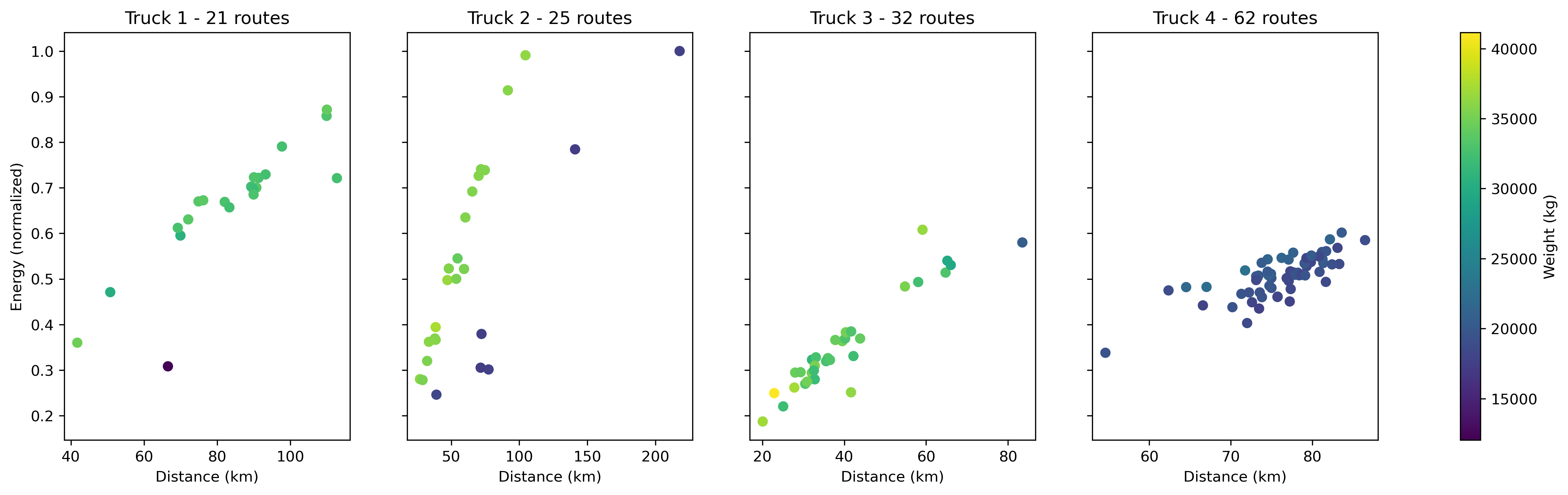}
\caption{Energy consumption vs distance for the individual trips with color-coded total weight.}
\label{fig:e_vs_d_w_m}
\end{figure*}

\begin{figure*}[ht]
\centering
\includegraphics[width=\textwidth]{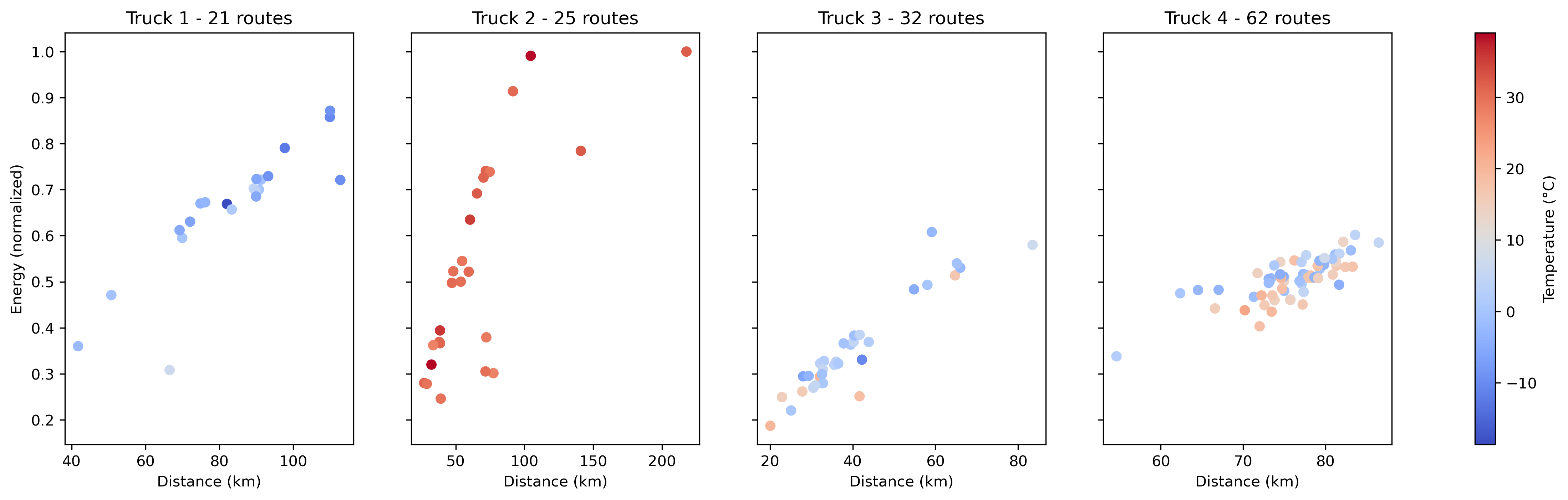}
\caption{Energy consumption vs distance for the individual trips with color-coded temperature.}
\label{fig:e_vs_d_w_t}
\end{figure*}

Figures \ref{fig:e_vs_d_w_m} and \ref{fig:e_vs_d_w_t} display information about the spread of the data in terms of propulsion energy consumed and distance traveled for each route with the different trucks. They also indicate how the relationships between the two differ with varying weight and temperature.

\bibliographystyle{plain}
\bibliography{mlep}

\end{document}